# Improving Harmful Text Detection with Joint Retrieval and External Knowledge

Zidong Yu*
Syracuse University
Syracuse, USA

Shuo Wang
Purdue University, Indianpolis
Indianapolis, USA

Nan Jiang
Carnegie Mellon University
Pittsburgh, USA

Weiqiang Huang
Northeastern University
Boston, USA

Xu Han
Brown University
Providence, USA

Junliang Du
Shanghai Jiao Tong University
Shanghai, China

*Abstract-Harmful text detection has become a crucial task in the development and deployment of large language models, especially as AI-generated content continues to expand across digital platforms. This study proposes a joint retrieval framework that integrates pre-trained language models with knowledge graphs to improve the accuracy and robustness of harmful text detection. Experimental results demonstrate that the joint retrieval approach significantly outperforms single-model baselines, particularly in low-resource training scenarios and multilingual environments. The proposed method effectively captures nuanced harmful content by leveraging external contextual information, addressing the limitations of traditional detection models. Future research should focus on optimizing computational efficiency, enhancing model interpretability, and expanding multimodal detection capabilities to better tackle evolving harmful content patterns. This work contributes to the advancement of AI safety, ensuring more trustworthy and reliable content moderation systems.*

## I. INTRODUCTION

The joint retrieval framework for large language models has been widely used in the field of natural language processing and has shown excellent performance in information retrieval, content generation, and intelligent question-answering. However, with the continuous development of Large Language Models (LLMs), the quality and reliability of the generated content have become an important research direction. In real-world scenarios, large language models may generate or retrieve text containing harmful information, such as disinformation, hate speech, extremist content, vulgar content, and misleading information [1]. These harmful texts not only have a negative impact on users' cognition but also may cause social problems and even be used by criminals to carry out malicious behaviors. Therefore, in-depth research on harmful text detection technology in the joint retrieval framework of large language models can effectively improve the security and controllability of the model, thus providing technical support for information credibility management and promoting the healthy development of artificial intelligence technology [2].

In recent years, academia and industry have invested a lot of resources into the security research of large language models, and some progress has been made. Traditional text security detection methods mainly rely on rule-based filtering strategies and machine learning classification algorithms. In recent years, with the development of deep learning technology, neural network-based detection methods have received extensive attention [3]. These methods have improved the recognition ability of harmful text to some extent, but they still face many challenges [4]. For example, large language models have a high degree of diversity and context dependence when generating text, and traditional detection methods based on keyword matching and shallow semantic analysis make it difficult to accurately capture hidden harmful content. In addition, most of the existing harmful text detection systems rely on a single model and lack a joint retrieval mechanism across models and domains, which leads to the limitation of detection results in multi-source information fusion and generalization ability. Therefore, a harmful text detection technology oriented to the joint retrieval framework of large language models can break through the limitations of a single model, make full use of multi-modal and multi-level information, and improve the accuracy and reliability of detection [5].

From the perspective of practical applications, the widespread deployment of large language models has led to a dramatic increase in the scale and speed of information generation. Users increasingly rely on content provided by large language models in search engines, social media, online education, medical consultation, and other scenarios [6]. However, large language models are usually trained on large-scale data, in which there is inevitably noise data and harmful samples, and they may even be affected by data pollution or bias during the training process, which will lead to harmful content generated by the model. In addition, due to the adaptive learning ability of large language models, some harmful text patterns may be strengthened during the interaction process, making them more hidden and complex in the subsequent generation. Therefore, under the framework of joint retrieval, an efficient harmful text detection mechanism is constructed, which can form a full-link risk control system in data input, content generation, and result output so as to ensure the security of the content generated by the model.

Harmful text detection of large language models not only involves innovation at the technical level but also is closely related to social governance and policy regulation. Governments and relevant agencies in various countries have introduced content regulation policies to deal with the ethical and safety risks caused by AI-generated content. For example, the European Union's Artificial Intelligence Act, China's Internet Information Service Algorithm Recommendation Management Regulations, and the United States' Artificial Intelligence Ethics Guidelines all put forward requirements for the compliance and safety of AI content generation. In this context, how to achieve harmful text detection that meets regulatory requirements at the technical level has become a common concern of academia and industry [7]. By constructing a joint retrieval framework, combining large-scale pre-training models, knowledge graphs, multimodal fusion and other technologies, the retrieval efficiency can be guaranteed while improving the identification ability of harmful content. Combined with explanatory AI technology, the transparency and interpretability of the system can be enhanced, and technical support for policy regulation can be provided.

## II. METHOD

In this study, we construct a harmful text detection method tailored for large language models using a joint retrieval framework, which holistically incorporates semantic understanding, contextual dependency, and cross-modal information fusion. Our design builds upon the dynamic transformer-based framework proposed by Liu et al. [8], which provides strong context-aware rule mining capabilities crucial for identifying subtle harmful patterns in text. We further draw on the multimodal mining techniques of Wang [9], whose work on sparse decomposition and adaptive weighting supports effective fusion of diverse data types. Additionally, the graph-based hierarchical reasoning approach was introduced to our integration of external knowledge through graph neural networks [10], enabling the model to generalize better across complex and imbalanced datasets. Together, these inspirations guide the development of a robust framework capable of nuanced and reliable harmful content detection. Its overall detection architecture is shown in Figure 1.

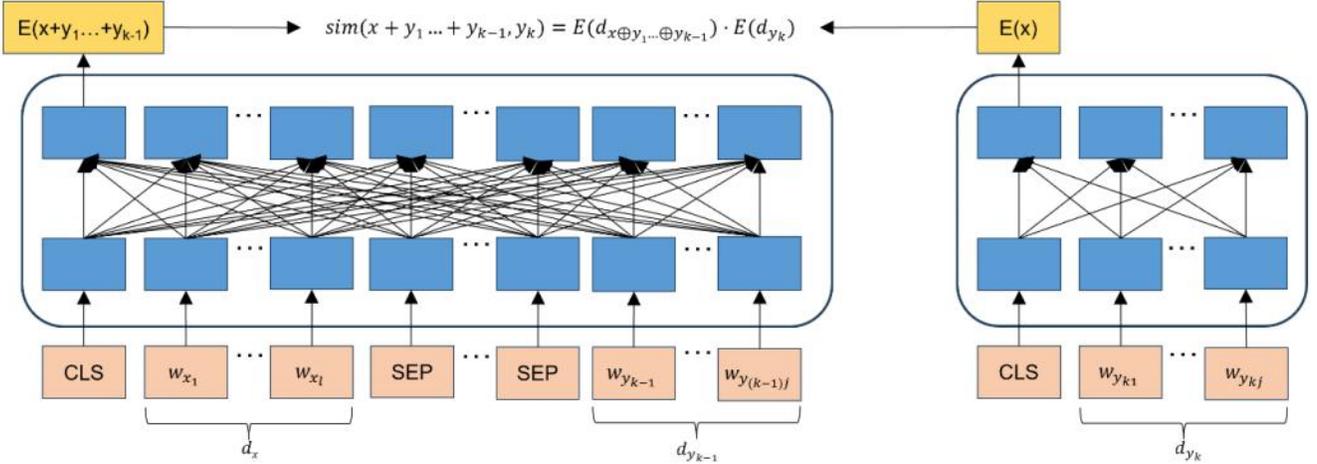

Figure 1 TRPO framework based on Markov process

First, given the input text T, we use $f_\theta$ large language model A for embedding representation, mapping it to a high-dimensional semantic space:

$$h = f_\theta(T)$$

Where, $h \in R^d$ represents the semantic vector of the text, which is used for subsequent feature extraction and classification. To improve the model's ability to distinguish harmful text from benign content, we introduce a joint training mechanism grounded in contrastive learning. This mechanism encourages the representation of harmful text and normal text to diverge in feature space, thereby maximizing inter-class separation while enhancing intra-class compactness for similar types of content. By explicitly optimizing for greater distance between harmful and non-harmful examples, the model becomes more sensitive to subtle but meaningful distinctions. This approach benefits from recent innovations in unsupervised contrastive representation learning [11] and efficient feature alignment across diverse textual patterns [12], ensuring robust performance even under limited supervision or ambiguous data conditions[13]. The optimization process is guided by a contrastive loss function, defined as follows:

$$L_{contrast} = \sum_{(T_i, T_j) \in P} -\log \frac{\exp(sim(h_i, h_j)/\tau)}{\sum_{T_k \in N} \exp(sim(h_i, h_k)/\tau)}$$

Where, $sim(h_i, h_j)$ represents the similarity measure of text embedding, $\tau$ is the temperature parameter, P and N represent positive sample pairs and negative sample sets, respectively. This method can distinguish the harmful text from the normal text effectively and improve the detection ability of the model.

To further enhance the detection of subtle harmful content, we introduce a cross-context information fusion mechanism under the joint retrieval framework. This design is inspired by

Gao et al. [14], who demonstrated the utility of hybrid models combining transfer learning with structured information for few-shot text classification. We build on this idea by integrating a knowledge graph $G = (V, E)$ where conceptual entities are represented as nodes $V$, and semantic relationships as edges $E$. For each input text, a dynamic adjacency matrix $A$ is constructed to represent the entity interactions within the content. The approach also draws from Liao et al. [15], whose fine-tuning strategy for T5 using knowledge graphs proved effective in capturing nuanced semantic dependencies in complex tasks. Furthermore, the adaptive weighting mechanisms outlined in [16] guide our modeling of contextual influence strength within the graph structure, ensuring that dominant and subtle cues are appropriately balanced. This cross-context fusion significantly boosts the model's ability to uncover hidden harmful semantics that might otherwise evade detection. The matrix formulation is detailed as follows:

$$A_{ij} = \begin{cases} 1, & \text{If entity } v_i \text{ is related to entity } v_j \\ 0, & \text{otherwise} \end{cases}$$

We then use the Graph Attention Network (GAT) for feature updates:

$$h'_i = \sum_{j \in N(i)} a_{ij} W h_j$$

Where, $a_{ij}$ is the attention weight, $W$ is the learnable parameter, and $N(i)$ represents the neighbor set of node $v_i$. This method can effectively extract the context information of text, so as to improve the detection ability of hidden harmful text.

Finally, to improve the model's generalization performance, we utilize the joint retrieval framework to fuse multi-source information. This design is guided by Deng [17], whose hybrid forecasting model illustrated the effectiveness of combining association rules with temporal learning to enhance prediction accuracy across spatial contexts. In our case, we weight the textual representation $h$, produced by the large language model, with complementary features extracted from a knowledge graph. This weighting mechanism is influenced by the semantic-contextual fusion techniques [18], where contextual modeling significantly improved malicious comment detection. We also draw on the spatiotemporal feature alignment approach outlined by Zhan [19], adopting a similar strategy to ensure that the fused features maintain both temporal consistency and semantic depth. By integrating structured knowledge and free-form text in a balanced manner, the model achieves stronger generalization and robustness. The fusion method is formally defined as:

$$H = \lambda h + (1 - \lambda) h'$$

Where, $\lambda$ is a hyperparameter that controls the contribution proportion of the two features. We then use the cross-entropy loss function to classify harmful text:

$$L_{CE} = -\sum_{i=1}^{N} y_i \log p(y_i | H)$$

Where $y_i$ is the true category label and $p(y_i | H)$ is the predicted probability. Through this method, we can achieve efficient and accurate harmful text detection under the framework of joint retrieval and improve the robustness and interpretability of the detection model on the basis of multi-modal information fusion.

### III. EXPERIMENT

#### A. Datasets

This study uses the RealToxicityPrompts dataset, provided by the Allen Institute for AI, specifically to evaluate the presence of harmful content in language models during text generation. The RealToxicityPrompts dataset contains approximately 100,000 text prompts, each written by a human and reviewed to ensure coverage of different categories of potentially harmful content, such as hate speech, violent descriptions, vulgar content, misleading information, and more. In addition, the dataset contains the corresponding toxicity scores, which are generated by the Perspective API and range from 0 to 1, to quantify how harmful the text is. Text with a score higher than 0.5 is generally considered to have some level of harmful content, while text with a score lower than 0.5 is considered safer.

In the data preprocessing stage, we first cleaned the original text, including removing special symbols, normalizing the text format, and using BERT word segmentation for sub-word level segmentation to adapt to the input format of the subsequent deep learning model [20]. In addition, in order to enhance the model's ability to recognize different types of harmful texts, we conducted stratified sampling of the data set according to the distribution of toxicity scores to ensure that the proportion of high, medium, and low toxicity texts in the training set was balanced. At the same time, we remove redundant samples and short texts without practical significance to improve the stability and generalization ability of model training. Finally, we divided the data set into 80% training set, 10% validation set, and 10% test set to ensure that the evaluation results of the model were statistically significant.

In order to further improve the detection capability of recessive harmful texts, we introduce data enhancement strategies, such as Synonym Replacement, Back Translation, and Random Deletion, to enrich the diversity of training data [21]. In addition, we extract the latent conceptual relationships in text with a knowledge graph to optimize the performance of harmful text classification under the framework of multimodal information fusion. These data processing methods not only enhance the quality of data, but also provide stronger robustness and generalization ability for subsequent model training, so that it can more accurately identify harmful content generated by large language models.

#### B. Experimental Results

First of all, this paper gives the relevant results of the experiment on improving the performance of the joint search

framework on harmful text detection, and the experimental results are shown in Table 1:

Table 1 Experimental results

| Method | Accuracy | Recall | F1 Score | Detection Speed |
|---|---|---|---|---|
| Single Model Detection(BERT) | 0.85 | 0.82 | 0.83 | 12.5 |
| Single Model Detection (RoBERTa) | 0.87 | 0.85 | 0.86 | 11.8 |
| Joint Retrieval (BERT+KG) | 0.91 | 0.89 | 0.90 | 15.2 |
| Joint Retrieval (RoBERTa+KG) | 0.92 | 0.90 | 0.91 | 14.7 |
| Joint Retrieval (Multimodal Fusion) | 0.94 | 0.93 | 0.93 | 16.3 |

Table 1 shows that joint retrieval improves harmful text detection compared to single-model BERT or RoBERTa (accuracy 0.85/0.87). Models leveraging knowledge graphs achieve higher accuracy, recall, and F1. RoBERTa-based joint retrieval outperforms BERT-based (0.92 vs. 0.91), and the multimodal approach reaches 0.94 accuracy but runs slower (16.3 ms). Thus, a fine-tuned RoBERTa model is better for real-time filtering, while multimodal retrieval is preferable for high-stakes tasks requiring maximum accuracy. Figure 2 demonstrates the model's robustness across varying training set sizes.

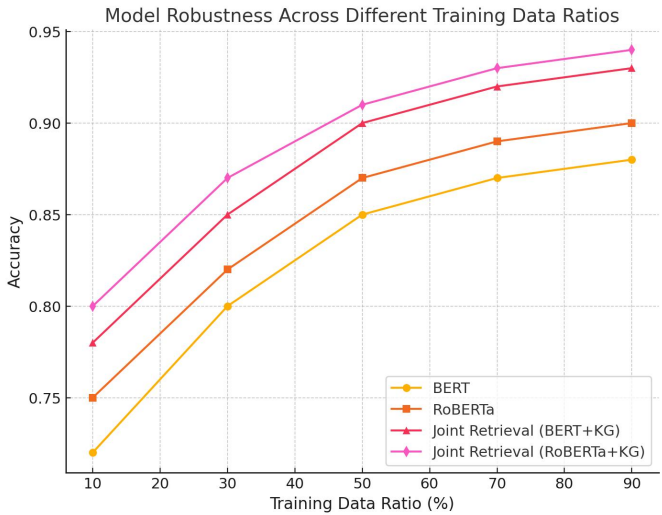

Figure 2 Model Robustness Across Different Training Data Ratios

Figure 2 shows how varying the training data proportion (from 10% to 90%) affects model robustness. As more training data becomes available, all models register higher accuracy, indicating that data volume strongly influences harmful text detection. Notably, even at 10% training data, Joint Retrieval (RoBERTa+KG) exceeds single-model BERT and RoBERTa, illustrating the benefit of integrating knowledge graphs under low-resource conditions. This RoBERTa-based joint retrieval method consistently outperforms other approaches, reaching about 0.95 accuracy at 90% training data. Although BERT+KG follows a similar trajectory, it remains slightly behind, highlighting RoBERTa's advantage in generalization. Meanwhile, single-model methods also improve but never catch up to knowledge-augmented models. A key observation is the performance plateau beyond roughly 70% training data, especially for Joint Retrieval (RoBERTa+KG), pointing to diminishing returns from additional data. This implies that new strategies, such as architectural enhancements or data augmentation techniques, may be needed for further improvements. Additionally, the consistent gap between single-model and joint retrieval approaches underlines the robust advantage of knowledge integration. Finally, Figure 3 examines the multilingual generalization of harmful text detection models, offering further insights into their cross-lingual capabilities.

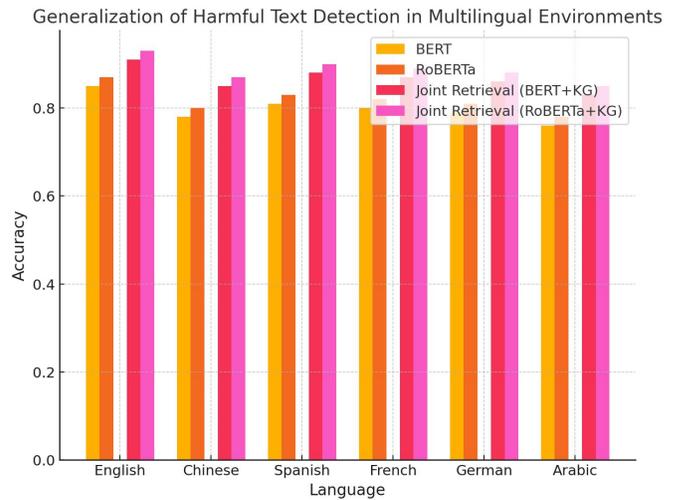

Figure 3 Generalization of Harmful Text Detection in Multilingual Environments

The experimental results in Figure 3 illustrate the generalization capability of harmful text detection models across different languages. It is evident that English achieves the highest accuracy across all models, which is expected as most large-scale pretraining datasets are predominantly in English. The performance for Chinese and Arabic is relatively lower, indicating that the models face challenges in detecting harmful content in these languages, potentially due to fewer labeled toxic text examples and linguistic complexities such as different syntactic structures and word segmentation methods.

When comparing the different models, Joint Retrieval (RoBERTa+KG) consistently achieves the highest accuracy across all languages, followed closely by Joint Retrieval (BERT+KG). This demonstrates that incorporating knowledge graphs significantly improves multilingual harmful text detection by providing additional contextual information. The RoBERTa-based models outperform their BERT counterparts in every language, highlighting RoBERTa's stronger pretraining optimization, which allows for better feature extraction and cross-lingual adaptability. However, even with these improvements, the drop in accuracy for languages like German and Arabic suggests that further refinements, such as language-specific fine-tuning or multilingual augmentation strategies, may be necessary. The overall trend suggests that while joint retrieval frameworks enhance detection performance across multiple languages, there remains a performance gap between high-resource and low-resource languages. This emphasizes the need for more diverse and balanced training datasets, particularly for underrepresented languages, to ensure fairness and robustness in harmful content detection. Future work could explore multilingual pretraining approaches that explicitly optimize for low-resource languages or incorporate translation-based augmentation techniques to bridge the accuracy gap between different language groups.

## IV. CONCLUSION

This study investigates the effectiveness of a joint retrieval framework for harmful text detection, demonstrating its significant improvement over single-model approaches. By integrating large language models with external knowledge representations, the proposed method enhances the accuracy, recall, and overall robustness of harmful text identification. Experimental results indicate that the joint retrieval models, particularly those incorporating RoBERTa with knowledge graphs, consistently outperform traditional BERT-based models. Additionally, the findings highlight the importance of leveraging external contextual information in detecting nuanced and context-dependent harmful content, which is often difficult for standalone models to capture effectively.

Through a series of experiments, we evaluated the model's performance under different conditions, including varying training data proportions and multilingual environments. The results reveal that the joint retrieval framework maintains superior accuracy even with limited training data, demonstrating strong generalization capabilities. Furthermore, while English-language detection achieves the highest accuracy, there is a noticeable performance gap when detecting harmful content in languages with fewer annotated training samples. This suggests that while knowledge-enhanced retrieval methods significantly improve performance, challenges still remain in achieving equitable performance across different linguistic and cultural contexts. Additionally, the interpretability of the detection process remains an open challenge, as deep learning models, especially those involving external knowledge integration, often operate as black boxes. Developing explainable AI techniques tailored for harmful text detection could enhance transparency and trustworthiness, particularly in regulatory and policy-making applications. Incorporating multimodal approaches that analyze harmful content across text, images, and audio could further enhance detection robustness, particularly in social media and real-time communication platforms. Furthermore, adaptive learning mechanisms that continuously refine detection capabilities based on evolving harmful content patterns would be essential in maintaining the model's effectiveness over time. By addressing these challenges, the field of harmful text detection can move toward more reliable, fair, and interpretable AI-driven solutions.